\documentclass[letterpaper, 10 pt, conference]{ieeeconf}  % Comment this line out if you need a4paper

\IEEEoverridecommandlockouts                              % This command is only needed if 
\usepackage{amsmath} % assumes amsmath package installed
\usepackage{amssymb}  % assumes amsmath package installed

\usepackage{comment}
\usepackage{multirow}
\usepackage{colortbl}
\usepackage{verbatim}
\usepackage{bm}
\usepackage{color}
\usepackage{graphicx}

\usepackage{longtable}
\usepackage{booktabs}

\title{\LARGE \bf
Probabilistic Human Motion Prediction via A Bayesian Neural Network}
\author{Jie Xu\textsuperscript{\textdagger}, Xingyu Chen\textsuperscript{\textdagger}, Xuguang Lan* and Nanning Zheng% <-this % stops a space
\thanks{{\textdagger} Equal contribution}
\thanks{* Corresponding author}
\thanks{J. Xu, X. Chen, X. Lan and N. Zheng are with the Institute of Artificial Intelligence and Robotics, Xi’an Jiaotong University, Xi’an
710049, China (e-mail: jie.xu@stu.xjtu.edu.cn; xingyuchen1990@gmail.com; xglan@mail.xjtu.edu.cn;
nnzheng@mail.xjtu.edu.cn)}% <-this % stops a space
}

\begin{document}

\maketitle
\thispagestyle{empty}
\pagestyle{empty}

%%%%%%%%%%%%%%%%%%%%%%%%%%%%%%%%%%%%%%%%%%%%%%%%%%%%%%%%%%%%%%%%%%%%%%%%%%%%%%%%
\begin{abstract}
Human motion prediction is an important and challenging topic that has promising prospects in efficient and safe human-robot-interaction systems. Currently, the majority of the human motion prediction algorithms are based on deterministic models, which may lead to risky decisions for robots. To solve this problem, we propose a probabilistic model for human motion prediction in this paper. The key idea of our approach is to extend the conventional deterministic motion prediction neural network to a Bayesian one. On one hand, our model could generate several future motions when given an observed motion sequence. On the other hand, by calculating the Epistemic Uncertainty and the Heteroscedastic Aleatoric Uncertainty, our model could tell the robot if the observation has been seen before and also give the optimal result among all possible predictions. We extensively validate our approach on a large scale benchmark dataset Human3.6m. The experiments show that our approach performs better than deterministic methods. We further evaluate our approach in a Human-Robot-Interaction (HRI) scenario. The experimental results show that our approach makes the interaction more efficient and safer.

\end{abstract}

%%%%%%%%%%%%%%%%%%%%%%%%%%%%%%%%%%%%%%%%%%%%%%%%%%%%%%%%%%%%%%%%%%%%%%%%%%%%%%%%
\section{INTRODUCTION}

Human motion prediction is an important and challenging topic in both computer vision and intelligent robotics communities which has been widely adopted to solve human intention understanding in human-robot-interaction scenarios. Supposing a robot is interacting with a human partner, for instance, handing a tool, playing games, or even taking care of the person, the interaction would be more efficient and safer if the robot could understand person's intention and accurately predicts person's future motion.

Currently, most of state-of-the-art human motion prediction algorithms \cite{Fragkiadaki2015,martinez2017human,Jain2016} rely on Recurrent Neural Networks (RNN). Given an observed motion sequence, these methods only predict a deterministic result, which brings a number of serious problems in the Human-Robot-Interaction (HRI) scenario. First, the robot should not make only one prediction since human motions are diverse in real interaction scenarios. The similar observed motions may correspond to very different future motions due to different intentions. Second, the robot does not know how much should it trust the predicted results. Specifically, if the observed motion has never been seen before, the predicted motion is actually untrustworthy. Third, the robot does not know how long should it trust the predicted motion. As RNNs could generate arbitrary length of human motions and the predicted results become more inaccurate as time goes, there are only a period of predicted motions trustworthy. Therefore, it is very risky for a robot to make decisions relying on a deterministic motion prediction algorithm.   

\begin{figure}[t!]
    \centering
    \includegraphics[width = 1\linewidth]{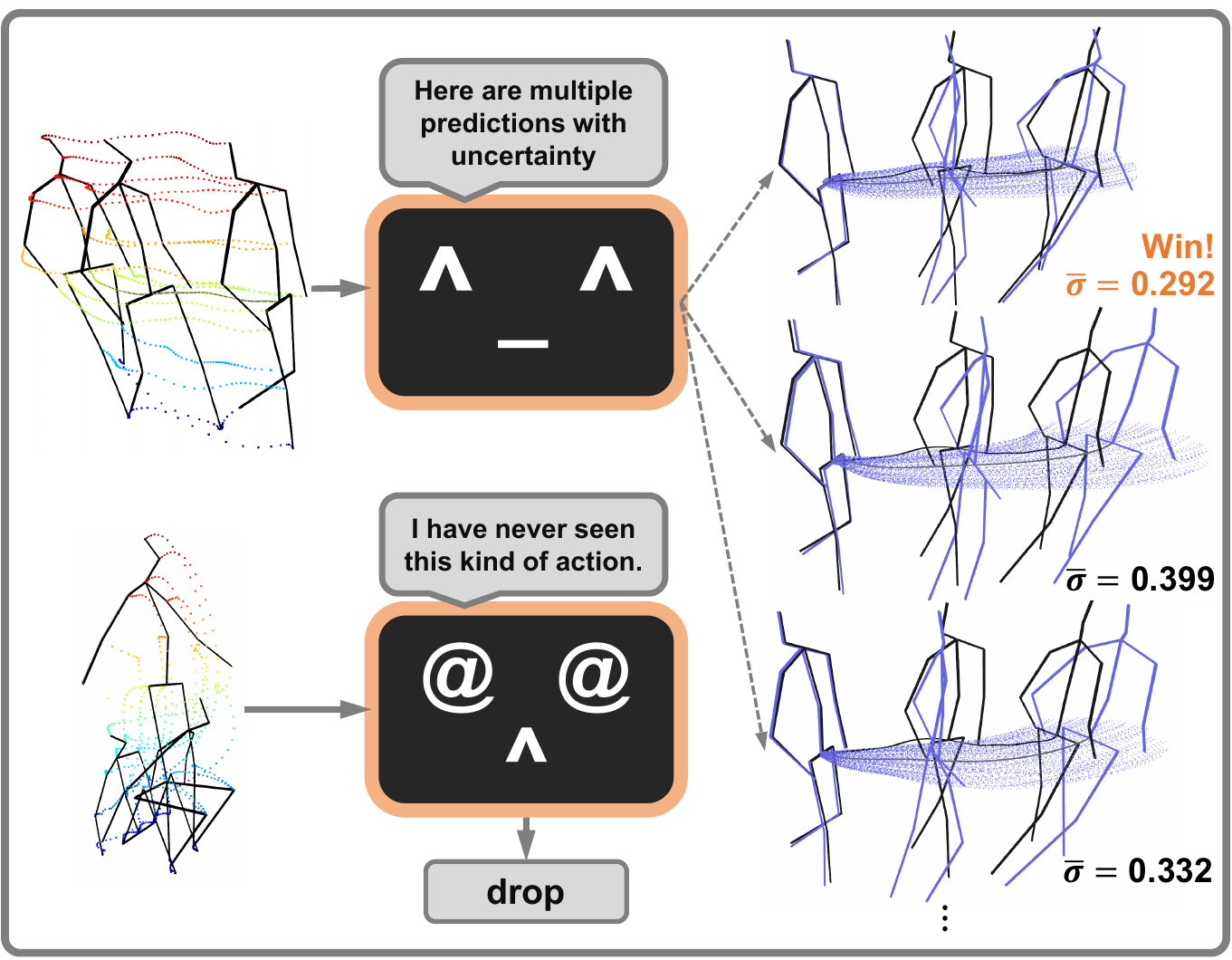}
    \caption{ A probabilistic human motion prediction algorithm should not only predict many possible results but also need to choose the optimal one to help a robot make the most appropriate decision. 
    %Our model can easily give a series of probabilistic predictions from a familiar motion, and then we choose the optimal prediction with the least uncertainty. The unfamiliar motion confuses the model, and the model discards all prediction results.
    }
    \label{fig:fig1}
\end{figure}

A more reliable human motion prediction algorithm in the human-robot-interaction system should be a probabilistic model. Previous methods such as \cite{barsoum2018hp,sidenbladh2002implicit} introduce Variational Auto-Encoders \cite{kingma2013auto} or Generative Adversarial Networks \cite{goodfellow2014generative} in the motion prediction algorithm which makes the model able to generate many possible results when given an observed motion. However, these methods could not further guide the robot to select the optimal prediction. In this situation, randomly select a predicted result is also risky. Thus, as illustrated in Fig.\ref{fig:fig1}, an ideal probabilistic model should not only predict many possible results but also need to tell the robot if it is risky to make predictions. Moreover it should give an optimal prediction to help the robot make the most appropriate decision. 

For this purpose, in this paper, we propose a novel probabilistic human motion prediction model based on a Bayesian Neural Network (BNN) \cite{mackay1992practical, neal1995bayesian}. The key idea of our approach is to extend the conventional deterministic motion prediction neural network to a Bayesian one. On one hand, as the BNN replaces the deterministic network’s weight parameters with distributions over these parameters, our model could predict a number of possible future motions when given an observed human motion sequence. On the other hand, the BNN based motion prediction model could naturally model two kinds of uncertainties simultaneously, including Epistemic Uncertainty (EU) and Aleatoric Uncertainty (AU) \cite{kendall2017uncertainties}. The EU captures the ignorance knowledge about the prediction model. Therefore, it can tell the robot whether an observed motion has been seen before. For the unseen observations, the robot could refuse to make risky decisions. The AU could captures noise inherent in the predicted motion sequences, enabling AU to reflect the confidence of the predicted results. Therefore, it helps the robot to select the most trustworthy parts of predicted results. 

In summary, our contributions in this paper are as follows:
\begin{itemize}
\item We propose a probabilistic approach for human motion prediction in the HRI scenarios. Given an observed motion sequence, the proposed approach not only predicts a number of possible future motions, but also tells the robot if the observation has been seen before. It also determines the optimal result among all possible predictions to guide the robot to make correct decisions.

\item We evaluate our approach on a popular human motion prediction benchmark dataset Human3.6m \cite{ h36m_pami} by designing a serials of experiments. The experimental results show that our approach performs better than deterministic motion prediction methods.
\item In a HRI scenario, we verify that the proposed approach makes the interaction more efficient and safer.
\end{itemize}

\section{Related work}
\subsection{Deterministic Human Motion prediction}

In recent years, with the development of deep learning technologies, there are a number of approaches \cite{Fragkiadaki2015,martinez2017human,Jain2016,NIPS2006_3078,butepage2017deep,gui2018adversarial,xu2019} which predict deterministic human motions by utilizing deep neural networks. For example, Encoder-Recurrent-Decoder (ERD) model \cite{Fragkiadaki2015} predicts the pose of human body based on the input videos. 
A Structural-RNN \cite{Jain2016} combines the high-level spatio-temporal graphs. It can convert the structural transformation from spatio-temporal map into RNN network using a hand-crafted graph. 
Modified from typical RNN structure, Martinez et al. \cite{martinez2017human} propose a sequence learning model with residual connection. Their network directly predicts the velocity of human motion and integrates the velocity into the pose of the previous frame. Given an observed motion, these methods only predict a deterministic result, which is usually problematic in HRI scenarios.

\subsection{Probabilistic Human Motion Prediction}
Most previous probabilistic human motion prediction use non-deep learning approaches \cite{NIPS2005_2783,Kuli2012,Lehrmann2014Efficient}. These methods usually rely on traditional machine learning algorithms such as Gaussian Process, Hidden Markov Model and Dynamic Forest Model. However, the crucial problem of these methods is that they cannot be applied to large scale datasets. Moreover, it is difficult for these methods to learn the variety of human motion patterns.
Recently, with the development of Variational Auto-Encoder \cite{kingma2013auto} and Generative Adversarial Networks \cite{goodfellow2014generative}, some probabilistic motion prediction approaches \cite{butepage2018anticipating,barsoum2018hp,sidenbladh2002implicit} which rely on deep neural networks, are proposed. For example, Butepage et al. utilize a conditional variational auto-encoder which makes the model able to generate many possible  results by sampling variables in a latent space. Barsoum et al. \cite{barsoum2018hp} propose an approach which learns a probability density function of future human poses conditioned on previous poses. It also could predict multiple sequences of possible future human poses. However, these methods could not be directly adopted in HRI scenarios since they could not further guide the robot to select the optimal prediction. Under this circumstance, randomly select a predicted result is also risky. Different from previous methods, we propose a novel probabilistic  human  motion  prediction model which based on BNN. Given an observed motion, our approach not only predicts a number of possible future motions, but also tells the robot if the observation has been seen before and also gives the optimal result among all possible predictions to guide the robot to make correct decisions.

\subsection{Bayesian Deep Learning}
Bayesian Neural Networks \cite{10.5555/3045118.3045290, 6790581, bui2015training, Okada_2020} usually replace the deterministic network’s weight parameters with distributions over these parameters, which gives network the ability to estimate the model uncertainty. Given an input, a BNN first samples weights from the weight distributions and then perform the forward calculation, which is always time-consuming. Gal \cite{Gal2016Uncertainty} finds that using Monte Carlo Dropout in a neural network during both training and testing could be approximately regarded as modeling  Bernoulli distribution over weight parameters. Based on the Dropout BNN, Kendall and Gal \cite{kendall2017uncertainties} propose a method to capture two kinds of uncertainties, including EU and HAU. Kendall~et~al.\cite{kendall2017uncertainties} propose a framework combining HAU (input-dependent) together with EU. A framework without retraining to calculate the model uncertainty is proposed by \cite{9001195}, however, which cannot distinguish Aleatoric Uncertainty (AU) and EU. 

% Park et al. \cite{ParkPM16a} use their intention-aware online planning algorithm to compute a reliable trajectory based on the predicted actions.

%% 在安全人体运动预测上，我们的方法通过使用贝叶斯深度学习同时捕获RZ不确定性和SJ不确定性，RZ不确定性用于检出unseen的动作序列，SJ不确定性用于感知预测误差。因此，我们的提出能够只预测有信心的序列，并给出安全误差范围内的最好概率预测。

%%%%%%%%%%%%%%%%%%%%%%%%%%%%%%%%% Method %%%%%%%%%%%%%%%%%%%%%%%%%%%%%%%%% 
\section{Method}

% 1 The similar observed motions may correspond to very different future motions due to different intentions. 
% 2 the robot does not know how much should it trust the predicted results.
% 3 the robot does not know how long should it trust the predicted motion.

%%%%%%%%% 任务引出，描述我们的approach %%%%%%%%
% 为了在真实环境中安全的预测可能的人体运动，我们给出了一个新颖的提出，它能给出可能的运动分布并规避不合理的预测（也许会导致危险）。本章我们将形式化的定义我们的方法。

\begin{figure*}[thpb]
    \centering
    %\framebox{
    \includegraphics[width = 1\linewidth]{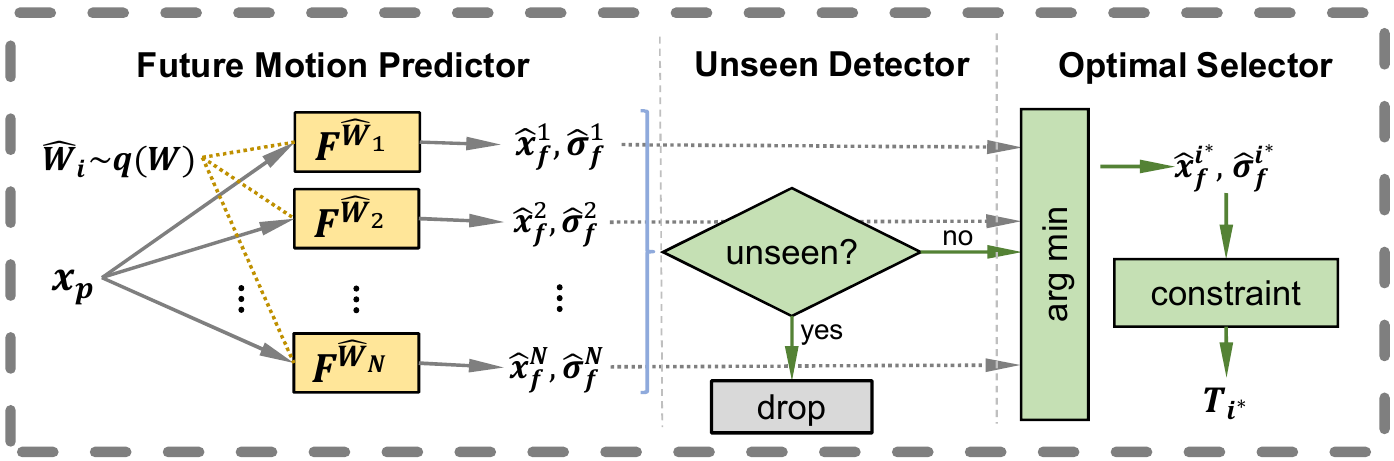}
    %}
    \caption{An overview of our model. Our model consists of three parts, including a Future Motion Predictor, an Unseen Motion Detector and an Optimal Motion Selector. 
    %The yellow boxes represent the Bayesian neural networks.
    }
    \label{fig:overview}
\end{figure*}

The goal of our approach is to make probabilistic and trustworthy human motion predictions for safe human-robot-interaction. In our approach, we represent human motion by a sequence of continuous 3D human skeleton poses. Each pose is represented as a vector that consists of the Cartesian coordinates of the joint locations.

Figure \ref{fig:overview} illustrates an overview of our model. It can be seen that our model consists of three modules, including a probabilistic future motion predictor, an unseen motion detector and an optimal motion selector. 
% When given a dataset $\mathcal{X}={X_1, ..., X_m}$, bayesian inference is used to compute the posterior $p(W|X)$ over the weights $W$.
% In Dropout variation inference techniques \cite{Gal2016Uncertainty}, the true posterior $P(W|X)$ is fitted with a simple distribution $q(W)$, which minimises the Kullback-Leibler(KL) divergence $D_{KL}(q(W)\|P(W|X))$.
When given an observed motion sequence $x_{-T_p:0}=\{x_{-T_p},...,x_{0}\}$ (abbreviated as $x_p$), our future motion predictor $F^{\hat{W}_i}$ with parameter $\hat{W}_i$ first predicts a number of probability results $(\hat{x}_{1:T_f}^i, \hat{\sigma}_{1:T_f}^i), i=1,...,N$(a pair of mean and standard deviation, abbreviated as $(\hat{x}_f^i, \hat{\sigma}_f^i)$), where $x_{t}=\{x_{t,1},...,x_{t,J}\}$, $J$ is  the  number  of  human  motion  joint, and $N$ is the number of samples, and the ground truth of future motion is $x_{1:T_f}$ (abbreviated as $x_{f}$), and $T_f$ is the maximum prediction length which is set to 2s.

Then, the unseen detector calculates EU of the predicted results to determine if the observed motion has been seen before. For an unseen input motion, EU would be large, which prevents the robot from making risky decisions. 
Finally, the optimal motion selector calculates HAU for each predicted future motion. Consequently, the optimal prediction $(\hat{x}_f^{i
^*},\hat{\sigma}_f^{i^*})$ and its trustworthy parts $(\hat{x}_{1:T_{i^*}}^{i^*},\hat{\sigma}_{1:T_{i^*}}^{i^*})$ could be identified, where $i^*$ is the index of the optimal prediction, and $T_{i^*}$ is the optimal length of its trustworthy parts.

\subsection{Future Motion Predictor}

\begin{figure}[t!]
    \centering
    \includegraphics[width=1\linewidth,height=4.5cm]{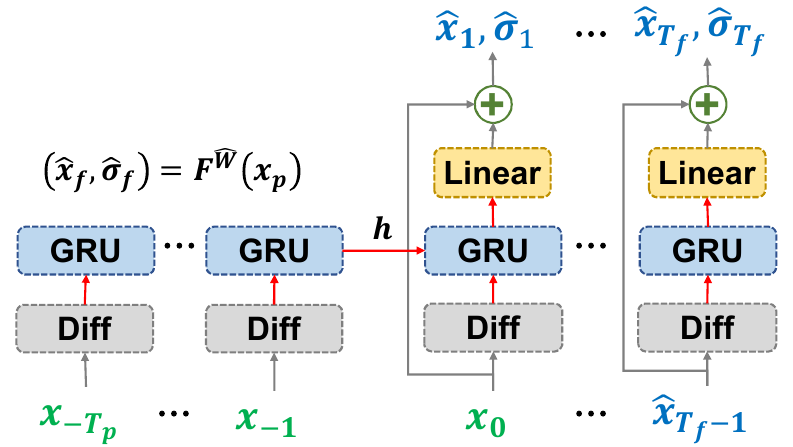}
    \caption{Architecture of the future motion predictor. The Monte Carlo Dropout is denoted by red arrows, the past motions are colored in the green, the predicted motions are colored in blue.}
    \label{fig:nn}
\end{figure}

%%%%%%%% 这一小节定义神经网络 %%%%%%%%

% figure
To generate probabilistic human future motions, we extend a previous deterministic motion prediction neural network Seq2Seq \cite{martinez2017human} to a Bayesian neural network which directly learns the distributions over the weight parameters $q(W)$. As a result, our future motion predictor can generate arbitrary number of possible results when given an observed motion sequence. Our model also consists of an encoder and a decoder. The encoder receives the input sequence $x_p$ and obtains a hidden variable $h$, and the decoder converts $h$ into possible future motions. If the network after dropout sampling $\hat{W}\sim q(W)$ is $F^{\hat{W}_i}$, our model can be described as $(\hat{x}_f^i,\hat{\sigma}_f^i)=F^{\hat{W}_i}(x_p)$. For convenience, we split $F$ into 2 functions:  $\hat{x}_f^i=f^{\hat{W}_i}(x_p)$ and $\hat{\sigma}_f^i= \sigma^{\hat{W}_i}(x_p)$.

% [todo] based on \cite{Li_2021}
The detailed network structure is shown in Fig. \ref{fig:nn}. Both encoder and decoder adopt Gated Recurrent Units (GRUs) \cite{Cho2014On} to recurrently make predictions over temporal domain. For the encoder, based on the survey results of \cite{Li_2021}, we first adopt a difference mechanism to capture the approximate velocity and acceleration of the current frame. Specifically, the acceleration can be obtained by the second-order difference of the previous three frames $x_{t-2:t}$. Further we take the current position $x_t$, velocity $\upsilon_t = x_t - x_{t-1}$ and acceleration $a_t = \upsilon_t - \upsilon_{t-1}$ as the human dynamics of the current frame. Finally, the human dynamics $[x_t, \upsilon_t, a_t]$ is fed to the GRU cell to calculate the hidden state $h$. 

For the decoder, the GRU predicts the speed $\hat{\upsilon}_t$ and the standard deviation of the position $\hat{\sigma}_t$ of current frame. The mean of the current position $x$ is obtained by adding $\hat{\upsilon}_t$ and the input position $\hat{x}_{t-1}$. Then the human dynamics $[x_t, \upsilon_t, a_t]$ is fed to the GRU cell for predicting the speed and the standard deviation of next frame. 

In our model, the Monte Carlo Dropout (MC-Dropout) is used for both training and testing.
On one hand, \cite{Gal2016Uncertainty} proves that Dropout is a reliable way to model the distributions over weight parameters. On the other hand, leveraging Dropout avoids directly sampling from weight parameter distributions, which is quite efficient for real time motion prediction systems.
It should be noted that the difference between MC-Dropout and naive Dropout is that the mask of MC-Dropout on each RNN time step is the same.
To simulate the unavailability of some key points caused by real sensors, our dropout unit takes the three coordinates of each key point as a whole. 

% 多插一句MCDropout

%============================ Framework ============================ % 

\subsection{Unseen Motion Detector}

% Epistemic

%%%%%%%% 引入我们方法中的认知不确定性 %%%%%%%%

% 贝叶斯神经网络和确定性神经网络的区别在于，贝叶斯神经网络的参数服从某一分布而不是固定的。
% 给定输入数据的情况下，认知不确定性的捕获是根据不同模型参数采样下输出的变化程度获得的。

Knowing whether the observed motion has never been seen before can prevent the robot from making risky decisions. As our motion predictor is a BNN, we can further design an unseen motion detector which leverages the EU to identify the unseen observations. Specifically, given an observed motion $x_p$, the unseen motion detector first calculate the EU $D(x_p)$ by Eq.(1),  
\begin{equation}
% (\hat{x}^i,(\hat{\sigma}^i)^2)
  D(x_p) = \frac{1}{N}\sum_{i=1}^{N}f^{\hat{W}_i}(x_p)^2 - (\frac{1}{N}\sum_{i=1}^{N}f^{\hat{W}_i}(x_p))^2,
\end{equation}
where $f^{\hat{W}_i}$ is our recurrent neural network with sampled weights using MC-Dropout. $f^{\hat{W}_i}(x_p)$ is the prediction results, $D(x_p)$ is obtained by subtracting the square of its mean (the right term) from its second-order moment of origin (the left term). In other words, $D(x_p)$ is a function of $x_p$, and $D(x_p)$ obtained by estimating the variance of $f^{\hat{W}_i}(x_p)$.

Then, the unseen motion could be easily identified by comparing the EU with a pre-defined threshold $D_\text{thd}$. Obviously, the threshold is very important which could directly affect the robot's decision-making. To find a proper threshold. We provide a simple yet effective way. The EU of each $x_m$ in the dataset $\mathcal{X}={x_1,x_2...,x_M}$ is ${D(x_m)}, m=1,2,...,M$. We sort $D(x_m)$ to $D_{(m)}$ in ascending order, and the threshold is given by $D_\text{thd}=D_{([0.95M])}$, where $[\cdot]$ is the rounding function. It means that 95\% of the data in the dataset would be accepted by the detector. $0.95$ is an empirical value that works well in most situations, because it can eliminate some noise data.
So, if $x_p$ satisfying $D(x_p)<D_\text{thd}$, it would be accepted, otherwise it would be rejected.

% \begin{equation}
%   R(x_p) = 
%   \left\{ 
%     \begin{array}{ll}
%       1, & D(x_p^m) \leq D_{thd} \\
%       0, & D(x_p^m) > D_{thd} \\
%     \end{array} 
%   \right.
% \end{equation}
% where 1 is set to accept and 0 to reject. 

%Detecting whether the input data is unseen is very important for security. The prediction of the model is based on the training data. In the case of no unseen detection, the predicted motion often makes unreasonable prediction.
%And we use epistemic uncertainty to realize unseen detection.

%In order to capture epistemic uncertainty, we model the network as Bayesian neural network by adding Monte Carlo Dropout \cite{Gal2016Uncertainty}. The parameters is $p(W|\mathcal{X})$, which is approximated by $q(W)$.
%In the previous work, epistemic uncertainty is given by the following formula:

%Now, we need to detect the input data $x_p$ is unseen or not according to the epistemic uncertainty. We define the epistemic uncertainty for $x_p^m$ is $D_m=D(x_p^m)$. After sorting in ascending order, we get $D_{(m)}, m=1,..,M$. The detecting function is defined as follow:

\subsection{Optimal Motion Selector}

%%%%%%%% 引入我们方法中的随机不确定性 %%%%%%%%
%%%%%%%%%%
For a motion sequence that has been seen before, the motion predictor generates $N$ possible results.
% $\{\hat{x}^{i}\}, i=1,2,...,N$
Then selecting the optimal prediction could help the robot make the most appropriate decision. For this purpose, we design the optimal motion selector which not only selects the best future motion $\hat{x}^{i^*}_f$ from all possible predicted results, but also determine the  maximal trustworthy length $T_{i^*}$ that satisfies some security conditions.

As our future motion predictor is built on top of GRUs which recurrently make predictions, the HAU naturally reflects the noisy scale of each predicted frame. Therefore, we can not only find the optimal one from all possible motions, but also drop the corrupted frames for selecting the trustworthy parts from the optimal motion.

To capture the HAU for each frame of the predicted motion sequences, as shown in Fig.\ref{fig:nn}, our model additionally output a standard deviation associated with the predicted human pose. The usual way is to modify the output dimension of the last layer of the network. Here we assume that the distribution of output is Gaussian $\mathcal{N}(\hat{x}^i_f, (\hat{\sigma}^i_f)^2)$. 
The index $i^*$ of the optimal prediction $(\hat{x}^{i^*}_f, \hat{\sigma}^{i^*}_f)$ of $x_p$ can be found by:
\begin{equation}
    i^* = \arg\min\limits_{i}\  \sum{\hat{\sigma}^i_f}
\end{equation}
% Then, the maximum $T_{i^*}$ can be found by:
% \begin{equation}
%     i^* = \arg\min\limits_{i}\  \sum{\sigma^{\hat{W}_i}(x_p)}
% \end{equation}

Then, if there is a maximum $T_{i^*}$ satisfying $\lambda\hat{\sigma}_{t}^i<e_{\max}$ for $t=1,2...,T_{i^*}$, then $T_{i^*}$ is the optimal length of the trustworthy parts. And we set the hyper-parameter $\lambda=1.28$.

% \begin{equation}
% \begin{aligned}
%   T_i =& \arg\max\limits_{\tau}\ 
%   {1:T_i}
% %   =& \arg\max\limits_{\tau}\ 
% %   \lambda_1 (\Phi(\hat{x}^i_{1:\tau} + e_{\max}) - \Phi(\hat{x}^i_{1:\tau} - e_{\max})) + \\
% %   &\lambda_2 (1 - \Phi{(\hat{x}^i_{\tau+1:T_f} + e_{\max})} + \Phi(\hat{x}^i_{\tau+1:T_f} - e_{\max}))
% \end{aligned}
% \end{equation}
% where $\Phi$ is the cumulative distribution function (CDF) of $\mathcal{N}(\hat{x}^i, (\hat{\sigma}^i)^2)$. 
%% 公式的第一项表明是我们要选一个预测长度使得选出的预测序列尽可能满足误差限制，
% The first term indicates that we have to choose a prediction length so that the selected prediction sequence meets the error limit. The second term is the trade-off term, which maximizes the length.

% 
% So the optimal prediction is $\hat{x}_{1:T_{\hat{i}}}^{\hat{i}}$, where $\hat{i}$ is given by:
% \begin{equation}
%     \hat{i} = \arg\max\limits_{i}\  \sum{\hat{\sigma}_{1:T_i}^i}
% \end{equation}

% where $\lambda_1, \lambda_2, \lambda_3, \lambda_4$ are hyperparameters.

% Now we can give the definition of the maximum length. When there is a maximum $T_{f_i}$ such that $2\hat{\sigma}_t < e_{\max}, t=1,...,T_{f_i}$ holds, then $T_{f_i}$ is called the maximum length of the sequence. Then the optimal motion is $\hat{x}_{\hat{i},1:T_{f_{\hat{i}}}}$, where $\hat{i} = \arg\max\limits_{i}(T_{f_i})$.

\subsection{Overall Objective Function}
%% 上一章中我们在输出放置了一个正态分布
Our model outputs the probability prediction at each time step. In the case of data independence, the objective function can be obtained by maximizing the posterior probability \cite{10.1023/A:1007665907178, 8578539}. We derive our loss function for BNN: %（具体方法)

\begin{equation}
    \mathcal{L} =
    \frac{1}{T_f}\frac{1}{J}\sum_{t=1}^{T_f}\sum_{j=1}^{J}\frac{\| x_{t,j} - \hat{x}_{t,j} \|^2}{2\hat{\sigma}_{t,j}^2} + \frac{1}{2}\log\hat{\sigma}_{t,j}^2
\end{equation}
The left term is weighted mean square error (MSE) between the predictions and ground truth in joint level, which is helpful for the variance to fit the prediction error from the data. The right term is a regularization term that helps to keep the variance at a reasonable value.
The weight decay term is omitted here.

\section{Experiments}
In this section, we first evaluate the proposed approach on a large scale benchmark dataset Human3.6m \cite{ h36m_pami}. We compare our approach with two baseline methods, including a state-of-the-art deterministic motion prediction model Seq2Seq \cite{martinez2017human} and a strong baseline Zero-velocity which takes the last input frame as all future predictions. Then we test our approach in a HRI  scenario to demonstrate its effectiveness. 

It worth noting that our Future Motion Predictor, Unseen Motion Detector and Optimal Motion Selector are denoted as FMP, UMD and OMS for short. We use ``+'' to indicate the combination of different parts. The numerical results of FMP and FMP+UMD are mean values since our approach could predict a number of possible results. The optimal numerical results are reported when we use the OMS.    

%The seq2seq \cite{martinez2017human} baseline is retrained under the above settings. The ``ours" method is the neural network without the unseen detector (d) and the optimal selector (s). We calculated Mean Per Joint Postion Error (MPJPE) between the predictions and the ground truth, as shown in the table

\subsection{Dataset}
The Human3.6m is currently the largest available motion capture dataset which is widely used for evaluating motion prediction algorithms. It contains 7 actors performing 15 varied activities such as walking, eating, smoking, discussion etc. In this dataset, each human pose is represented by a high dimensional vector of 3D joint locations.

\subsection{Implementation Details}
All our experiments are conducted on Pytorch 1.4 \cite{paszke2017automatic}. The encoder of the FMP consist of a single GRU cell with 1440 neurons. The decoder consists of a GRU cell and a Fully Connected layer with 1440 and 51 neurons. We adopt the Adam optimizer \cite{Kingma2014Adam} to train our model. The learning rate is set to $0.0005$ and the weights decay rate is set to $0.0001$. The batch size is set to $128$. For each Dropout layer, the probability parameter is set to $50\%$. Our visualization results are rendered using matplotlib and open3d \cite{Hunter2007}.

% 本章，我们在Human3.6m数据集和我们自己的数据集上评价了我们的方法。
%The proposed method is evaluated using the real data (partly unseen). And large dataset HUMAN 3.6M \cite{IonescuSminchisescu11, h36m_pami} and our own small dataset is used. 
% 实验部分我们详细评估了我们的方法在面对真实数据（部分unseen）时的表现。
% In this section, we evaluated the performance of our method using the real data (partly unseen) in detail.
%The public dataset Human 3.6M is currently one of the largest available motion capture datasets. It contains 7 subjects, and each subject contains 15 actions (walking, eating, smoking, discussion, directions, greeting, phoning, posing, purchases, sitting, sittingdown, takingphoto, waiting, walkingdog, walkingtogether). 
%Our small dataset contains 2 actions (walking, reaching), which consists of 30 minutes at all. It is captured using Kinect DK.
% (walking, eating, smoking, discussion, directions, greeting, phoning, posing, purchases, sitting, sittingdown, takingphoto, waiting, walkingdog, walkingtogether).
% 对应
% 

% \subsection{Parameter Settings}
%In our method, the experiment platform is pytorch 1.4.
%We use a single GRU cell with 1440 units. The ADAM optimizer \cite{Kingma2014Adam} is used with weights decay rate 0.0001 during training, and the learning rate is set as 0.0005. 
%The batch size is set to 128. We feed 48 frames of motion to model in 25fps. All dropout probability is $0.5$, and $N=64$ for all sampling.

\subsection{Unseen Motion Detector}

\begin{table*}[htbp]
  \centering
  \footnotesize
  \renewcommand\arraystretch{0.8}
  \caption{The MPJPE on acceptance at different time steps and unseen samples detection rate are reported for different methods. The best results are highlighted by bold numbers.}
\setlength{\tabcolsep}{4.8mm}{
    \begin{tabular}{ccccccccc}
    \toprule
    \multirow{2}[4]{*}{model} & \multicolumn{1}{c}{\multirow{2}[4]{*}{Train Set}} & \multicolumn{1}{c}{\multirow{2}[4]{*}{Test Set}} & \multicolumn{1}{c}{\multirow{2}[4]{*}{Det. \%}} & \multicolumn{5}{c}{MPJPE on acceptance} \\
\cmidrule{5-9}          &       &       &       & 400ms & 800ms & 1200ms & 1600ms & 2000ms \\
    \midrule
    Zerovel. & -     & walking & -     & 0.111 & 0.161 & 0.173 & 0.201 & 0.249 \\
    Seq2Seq & walking & walking & -     & 0.056 & 0.066 & 0.073 & 0.080 & 0.087 \\
    FMP  & walking & walking & -     & 0.046 & 0.060 & 0.069 & 0.077 & 0.088 \\
    FMP+UMD & walking & walking & 4.82\% & \textbf{0.045} & \textbf{0.059} & \textbf{0.068} & \textbf{0.075} & \textbf{0.086} \\
    \midrule
    Zerovel. & -     & eating & -     & 0.053 & 0.084 & 0.102 & 0.116 & 0.137 \\
    Seq2Seq & walking & eating & -     & 0.149 & 0.204 & 0.242 & 0.248 & 0.245 \\
    FMP   & walking & eating & -     & 0.086 & 0.135 & 0.163 & 0.184 & 0.209 \\
    FMP+UMD & walking & eating & 67.65\% & \textbf{0.048} & \textbf{0.072} & \textbf{0.089} & \textbf{0.100} & \textbf{0.112} \\
    \midrule
    Zerovel. & -     & sittingdown & -     & 0.068 & 0.116 & 0.152 & 0.181 & 0.205 \\
    Seq2Seq & walking & sittingdown & -     & 0.298 & 0.400 & 0.461 & 0.490 & 0.514 \\
    FMP   & walking & sittingdown & -     & 0.166 & 0.251 & 0.289 & 0.323 & 0.352 \\
    FMP+UMD & walking & sittingdown & 100.00\% & -     & -     & -     & -     & - \\
    \bottomrule
    \end{tabular}%
    }
  \label{tab:exp2}%
\end{table*}%

We first conduct experiments to demonstrate how the unseen detector affects the predicted results on Human3.6m dataset. In this experiment, all methods are trained on action ``walking", and then evaluated on three different actions, including ``walking", ``eating" and ``sittingdown". We compare our approach with the baseline methods using two metrics including the unseen samples Detection rate over and Mean Per Joint Position Error (MPJPE) on acceptance in terms of meters. The results are reported in Table \ref{tab:exp2}. The constant velocity \cite{scholler2020constant} is not included in the table because of its serious deformation predictions on human motion prediction.

First, we see that only with the help of UMD, the unseen test sample would be detected. There are $4.82\%$ of the test samples in ``walking" which have a large EU so that they are regarded as unseen samples. It is because the data distributions of the train set and the test set are not exactly the same. For these samples, our FMP has a lower confidence for making predictions. On ``eating" and ``sittingdown" test sets, $67.65\%$ and $100\%$ of the test samples are detected as unseen. The reason is that the action ``eating" includes eating with sitting, eating with standing, eating with walking, which overlaps with the action ``walking" partly. The action ``sittingdown" includes sitting on a chair, sitting on the ground, etc. There is almost no overlap with ``walking".

\begin{figure}[t]
    \centering
    \includegraphics[width = 1 \linewidth]{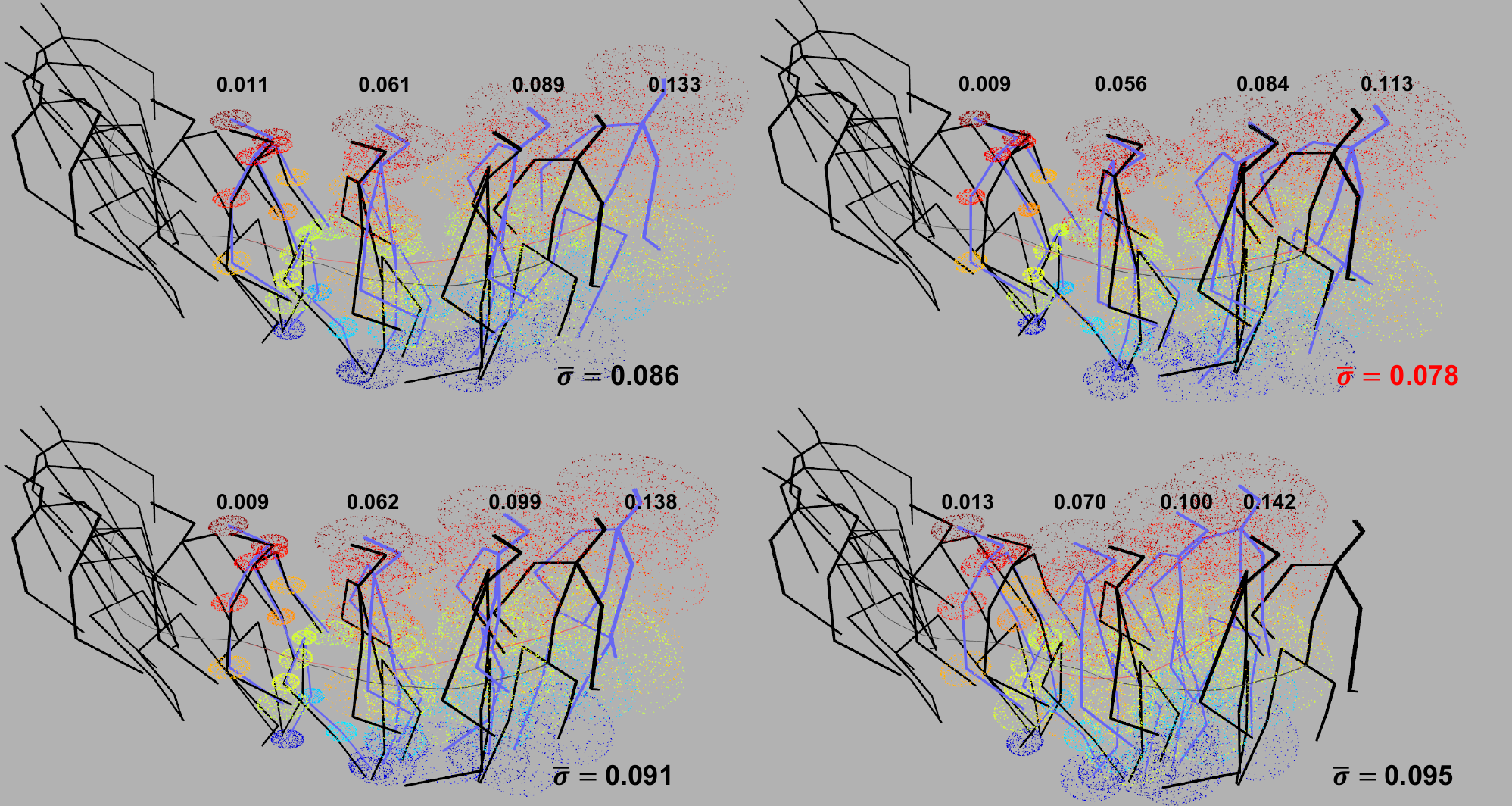}
    \caption{Visualization results of the predicted future motions. The average $\bar{\sigma}$ is annotated above each frame. The optimal result has the smallest average  $\bar{\sigma}$. }
    \label{fig:exp3_2}
\end{figure}

\begin{figure}[t]
    \centering
    \includegraphics[width = 1\linewidth]{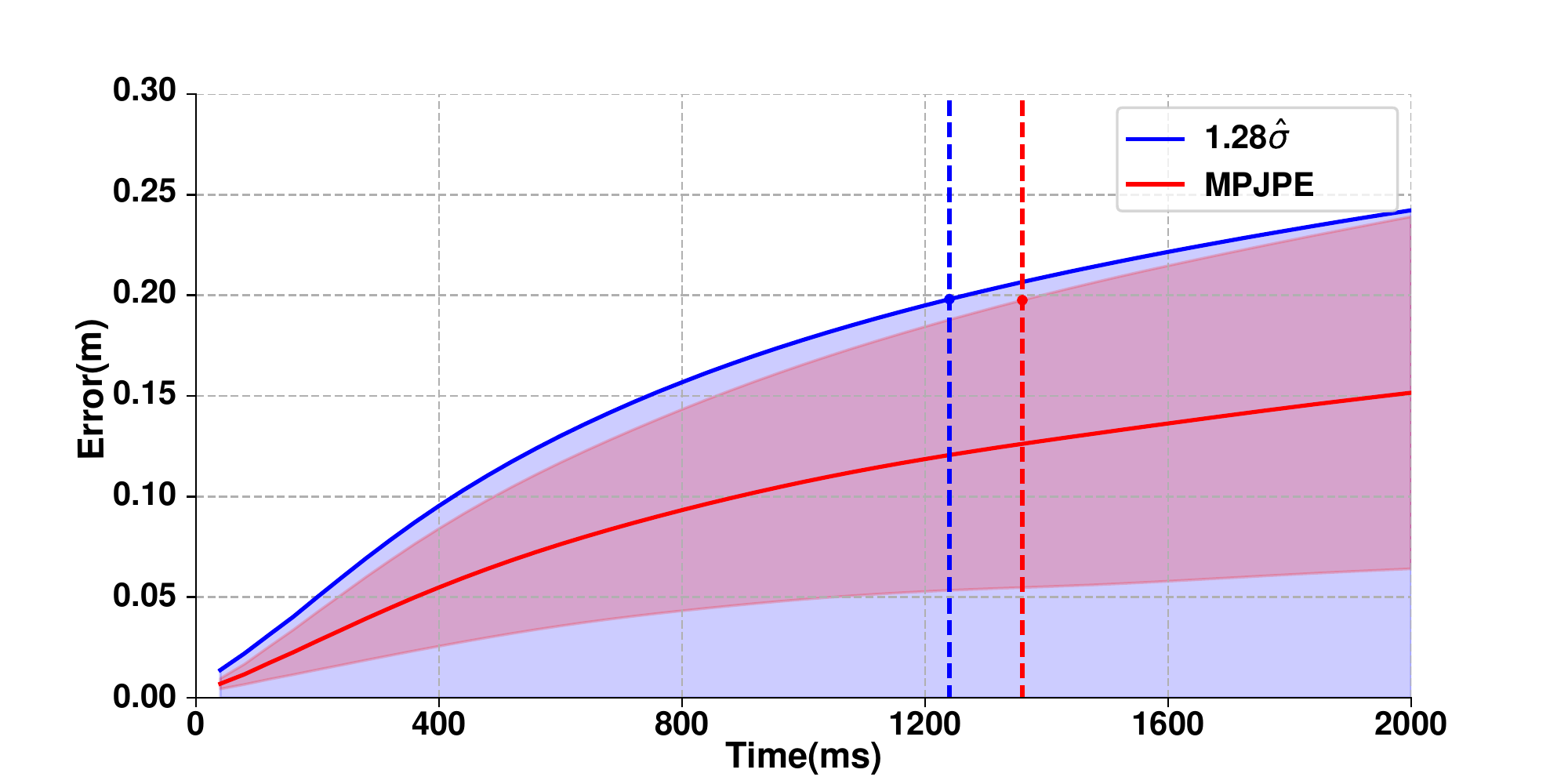}
    \caption{The real maximal trustworthy length (red dash line) is calculated by using ground-truth motions. The predicted maximal trustworthy length (blue dash line) is calculated by using HAUs.}
    \label{fig:exp3}
\end{figure}

Second, on the ``walking" test set, our FMP achieves better MPJPE results compared with the baselines. It shows that training the motion predictor with the constrain of uncertainty could also improve its performance. on the ``eating" and ``sittingdown" test sets, both Seq2Seq and FMP are worse than Zerovel., which indicates that making prediction for unseen motion sequences is even worse than predicting the future motion as a static pose.

Third, it can be seen that combining the FMP with UMD achieves the lowest MPJPE results on ``walking" and ``eating". It refuses to make predictions for ``sittingdown" since all test samples are regarded as unseen. In other words, using the UMD could effectively identify the unseen motion sequences to prevent FMP from making bad predictions.

\begin{figure*}[thpb]
    \centering
    \includegraphics[width = 1\linewidth, height=0.26\linewidth]{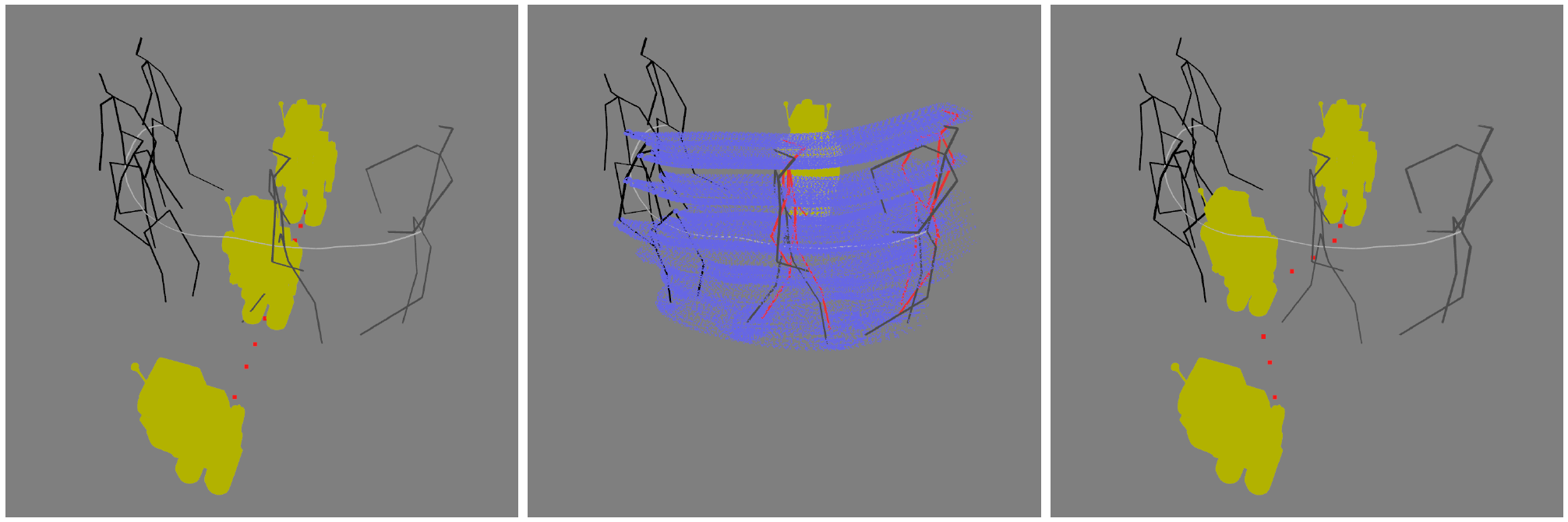}
    \caption{Trajectory visualization results of the robot in a HRI scenario. Each figure shows the positions of the human and the robot at different time steps.The left figure shows the robot's trajectory without making prediction and optimization. The middle figure shows the predicted human motions after the robot observes the human for 2s (red indicates the mean value of the prediction, and the blue scatter represents the 95\% confidence interval). The right figure shows the optimized robot's trajectory when taking the predicted human motions into consideration.}
    \label{fig:exp4}
\end{figure*}
\subsection{Optimal Motion Selector}
We further conduct experiments to intuitively demonstrate the effectiveness of the Optimal Motion Selector in this section. 

Given an observed motion sequence, our FMP predicts a number of possible future motions, in which each frame is attached with a predicted $\hat{\sigma}$. By simply calculate the mean $\overline{\sigma}$ for each motion, the OMS finds the optimal one. Fig. \ref{fig:exp3_2} shows the visualization results of the predicted motions which correspond to a same observed motion. It can be seen that the optimal result, which has the smallest $\overline{\sigma}=0.078$, is most close to the ground-truth motion.

Fig. \ref{fig:exp3} shows our OMS also could accurately predict the maximal trustworthy length according to the HAUs. In this experiment, we set the error threshold to $0.20$m. In other words, we think the predicted frames whose MPJPE is larger than $0.20$m need to be abandoned. By comparing the predicted results with the ground-truth motions, we get the real maximal trustworthy length which equals to $1360$ms. By using the predicted HAUs, OMS predicts the maximal trustworthy length as $1240$ms, which is very close to the real maximal trustworthy length. It worth noting that the real maximal trustworthy length could hardly be achieved in real time HRI scenarios since the ground-truth motions are usually unavailable. In this situation, OMS provides a reliable solution. 

We present the numerical results to show the effectiveness of OMS in the next section.

%This experiment evaluates whether the optimal prediction $\hat{x}_{1:T_{\hat{i}}}^{\hat{i}}$ given by our method meets the error limit. The evaluation protocol uses PCK3D, and the purpose is to evaluate the proportion of joint points that meet the error limit in the given prediction.
%It should be noted that in the two methods of zerovel and seq2seq, we intercept the prediction with length $T_f=50$ for evaluation. And in the "ours d" method, we average the $N$ predictions for evaluation.

% $T_{opt}$ guarantees the error limitation (e.g. 0.20m) calculated by $1.28\hat{\sigma}$ or MPJEP is shown with the blue or red dash line.

%\begin{figure}[thpb]
%    \centering
%    \includegraphics[width = 1.\linewidth]{graph/realhri.jpg}
%    \caption{}
%    \label{fig:exp4_real}
%\end{figure}

% \begin{figure}[thpb]
%     \centering
%     \includegraphics[scale=0.44]{graph/exp3.pdf}
%     \caption{}
%     \label{fig:exp3}
% \end{figure}

% 对于安全人机交互来说，低估的误差往往会产生严重的后果，但是过于高估又会导致。图展示了不同置信区间下低估的百分比，

% 平均标准差与平均MPJPE
% 如图所示，预测的标准差平均上来说大于预测的MPJPE，也就是说我们通常会高估误差。红线是我们根据预测的标准差在误差限制内选取的最优安全长度。
% 图中显示了

%% 还需介绍评价指标 但是评价指标还有点问题，还得思考一下
% 可能用图表会更好一点

\subsection{Future Motion Prediction}

\begin{table}[t]
  \centering
  \caption{The MPJPE on acceptance at different time steps are reported for different methods.}
  \resizebox{1.0\linewidth}{!}{
    \begin{tabular}{cccccc}
    \toprule
    method  & 400ms   & 800ms   & 1200ms  & 1600ms  & 2000ms \\
    \midrule
    Zerovel.   & 0.072 & 0.113 & 0.138 & 0.158 & 0.179 \\
    Seq2Seq     & 0.063 & 0.099 & 0.125 & 0.144 & 0.160 \\
    FMP     & 0.057 & 0.097 & 0.123 & 0.141 & 0.156 \\
    FMP+UMD  & 0.055 & \textbf{0.093} & \textbf{0.119} & 0.136 & 0.152 \\
    FMP+UMD+OMS  & \textbf{0.054} & \textbf{0.093} & \textbf{0.119} & \textbf{0.135} & \textbf{0.151} \\
    \bottomrule
    \end{tabular}}
  \label{tab:overall}%
\end{table}%

In this experiment, we compare our approach with the baseline methods on all 15 actions in the Human3.6m dataset. The MPJPE results are reported in Table \ref{tab:overall}. First, we see that FMP performs better than the two baseline method since it achieves a lower MPJPE at each time step. This is consistent with our previous conclusion that training the motion predictor with the constrain of uncertainty could improve the predictor's performance. Second, with the help of UMD, there are $4.30\%$ of the observed motion sequences are regarded as with a high EU which are directly dropped without making any prediction. Therefore, the MPJPE on acceptance of each time step is further decreased. Third, when we continue to introduce the OMS, our approach achieves the best results, which indicates the OMS could select the optimal future motion among all possible motions. Here, the OMS only finds the optimal predicted motion without dropping any frames with high HAU.

\subsection{Safe Robot Trajectory Optimization}

% 目的
% 这个实验的目的是在人机协作环境中评估我们的预测能多大能力够帮助机器人避免和人类碰撞。
% 设置
% 
In this experiment, we test our approach in a HRI scenario. Specifically, we demonstrate whether our approach could help the robot avoid collisions when interacting with a human. Inspired by CHOMP \cite{5152817}, we define our trajectory optimization objective function by replacing the original distances with the predicted uncertainty. 

Fig. \ref{fig:exp4} visualizes the trajectories of the robot and human in a HRI scenario. In the left figure, we see that the robot collided with the human, since the robot did not predict the human's motion. The middle figure shows the predicted human motions after the robot observes the human for 2s. Then, as shown in the right figure, by taking the future human motion into consideration, the robot optimizes its trajectory for avoiding collision.

\section{Conclusion}

In this paper, we propose a probabilistic approach for human motion prediction in human-robot-interaction scenarios. The proposed is based on a Bayesian neural network which aims to learn the distributions over weight parameters. On one hand, it is capable of predicting a number of possible future motions when given an observed human motion sequence. On the other hand, it could capture two kinds of uncertainties simultaneously to tell the robot whether the predicted results are trustworthy. In this way, it helps the robot to make the most appropriate decision. We first validate our approach on a benchmark dataset and then evaluate it in a HRI scenario. The experimental results demonstrate the effectiveness of our approach.

\section*{ACKNOWLEDGMENT}
This work was supported in part by NSFC under grant No.62088102, No.91748208, NSFC No.61973246, Shaanxi Project under grant No.2018ZDCXLGY0607, and the program of the Ministry of Education.

% This work was supported in part by Trico-Robot plan of NSFC under grant No.91748208, National Major Project under grant No.2018ZX01028-101, Shaanxi Project under grant No.2018ZDCXLGY0607, NSFC under grant No.61973246, and the program of the Ministry of Education for the university.

{
\bibliographystyle{IEEEtran}
\bibliography{egbib}
}
\begin{comment}

\end{comment}

\end{document}